\documentclass[conference]{IEEEtran}
\IEEEoverridecommandlockouts

\usepackage{url}

\usepackage{tabularx}
\usepackage{booktabs}
\usepackage{amsmath, amssymb}
\usepackage{mathtools}
\usepackage{graphicx}
\usepackage{stmaryrd}
\usepackage{multirow}
\usepackage{xcolor}

\DeclareMathOperator{\norm}{\mathcal{N}}

\DeclareMathOperator{\unift}{Unif_2}

\DeclareMathOperator{\logit}{logit}
\DeclareMathOperator{\probit}{\Phi^{-1}}

\newcommand{\bcov}{\rho_{ij}}
\newcommand{\ncov}{\tau_{ij}}

\newcolumntype{Y}{>{\raggedleft\arraybackslash}X}
\newcolumntype{Z}{>{\centering\arraybackslash}X}

\title{Estimating Conditional Covariance between labels for Multilabel Data}

\author{\IEEEauthorblockN{Laurence A.~F.~Park}
\IEEEauthorblockA{School of Computer, Data and Mathematical Sciences\\
Western Sydney University, Australia\\
Email: l.park@westernsydney.edu.au}
\and
\IEEEauthorblockN{Jesse Read}
\IEEEauthorblockA{LIX, Ecole Polytechnique\\
Institut Polytechnique de Paris, France\\
Email: jesse.read@polytechnique.edu}}

\begin{document}

\maketitle

\begin{abstract}
  Multilabel data should be analysed for label dependence before
  applying multilabel models.  Independence between multilabel data
  labels cannot be measured directly from the label values due to
  their dependence on the set of covariates $\vec{x}$, but can be
  measured by examining the conditional label covariance using a
  multivariate Probit model. Unfortunately, the multivariate Probit model
  provides an estimate of its copula covariance, and so might not be
  reliable in estimating constant covariance and dependent covariance.
  In this article, we compare three models (Multivariate Probit,
  Multivariate Bernoulli and Staged Logit) for estimating the
  constant and dependent multilabel conditional label covariance.  We
  provide an experiment that allows us to observe each model's
  measurement of conditional covariance. We found that all models
  measure constant and dependent covariance equally well, depending on
  the strength of the covariance, but the models all falsely detect
  that dependent covariance is present for data where constant
  covariance is present. Of the three models, the Multivariate Probit
  model had the lowest error rate.
\end{abstract}

\section{Introduction}

Multilabel data consisting of a set of labels $\vec{y} \in \{0,1\}^L$
with covariates $\vec{x} \in \mathbb{R}^M$ appears in many domains of
science (e.g.
\cite{swaminathan2024multi,ozturk2025hydravit,gao2024mineral}).
Each label $y_i$ in a multilabel data set is modelled as a function of the
covariates $\vec{x}$, where the multilabel model provides the
additional precision by also modelling the dependence between labels.
If there is no dependence between some of the labels, the data can be
split and modelled separately; if there is no dependence between all
labels each label can be modelled separately using a binary
classifier, providing a simpler and faster analysis.
Therefore, before fitting a multilabel model, the multilabel data
should be examined to determine the relationship between labels.

It is a common misconception that a multilabel model is required if
there is dependence between the labels, when in fact all labels will
be dependent due to each label being a function of $\vec{x}$. Therefore,
instead of checking for independence, the conditional independence of each label $y_i$ conditioned on $\vec{x}$
is required \cite{dembczynski2012label}.
Previous studies have investigated the utility of the multivariate Probit
model for assessing conditional independence of labels
\cite{liu2019copula}. Unfortunately, initial investigations
reveal that the multivariate Probit model is not measuring the label
dependence directly \cite{park2025pakddAnon}.

In this article, we evaluate the utility of three analytical models for assessing
the conditional covariance of labels in multilabel data (Multivariate Probit,
Multivariate Bernoulli and Staged Logit). The models are presented and a
controlled experiment is run to identify the ability of each model to
detect conditional covariance between labels.
The contributions of this article are:
\begin{itemize}
\item A detailed description of the problem when using the multivariate Probit model for detecting label dependence (Section \ref{sec:normprob}).
\item Presentation of two alternative models for assessing label dependence (Section \ref{sec:anternative}).
\item An experiment assessing each model's ability to assess label dependence (Section \ref{sec:experiment}).
\item An examination of each model's approximations (Section \ref{sec:analysis}).
\end{itemize}
Note that we are not proposing a new multilabel classification
model. We are investigating the use of analytical models for detecting
conditional covariance in multilabel data.

The article will proceed as follows: Section \ref{sec:back}
introduces multilabel classification and the need to measure
conditional covariance in multilabel data,
Section \ref{sec:probit}
describes the multivariate Probit model and its problems. Section
\ref{sec:anternative} presents two alternative models for measuring
label dependence. Section \ref{sec:experiment} provides an experiment
and the results from the three models.
Finally, Section \ref{sec:analysis} examines the models to provide
insight into the experimental results.

\section{Background}
\label{sec:back}

Multilabel classification is a generalisation of Binary classification
from one response to many responses, called labels. A multilabel classifier takes
input vector $\vec{x} \in \mathbb{R}$ and provides an associated
set of $L$ binary predictions $\vec{y} \in \{0,1\}^L$, one for each of
the $L$ labels. For example,
the input data could be a vector representing an retinal image and the
output be the absence or presence of a set of $L$ diseases detected
from the image.

A multilabel classifier has a label space of size $2^L$ (the number of
possible label states) and therefore can be mapped to a multiclass
classification problem predicting one of the $2^L$ possible states.
As $L$ increases, the label space grows exponentially; for example for
50 labels, there are $1.1259\times 10^{15}$ possible solutions making
the multilabel classification task difficult and time consuming when
$L$ is large.

If we assume that all of the labels are independent, a multilabel
classification problem over $L$ labels can be split into a set of $L$
binary classifiers. If the labels are not independent, we can take
advantage of the dependence to improve the classification accuracy,
but the complexity of the model increases.  Numerous methods have been
proposed to deal with the large label space
\cite{read2009classifier,LEAD,liu2020synthetic,lanchantin2020neural,ExtML2,mylonas2023persistence}.
For example, label powerset subset methods \cite{RAkEL2,MetaLabels},
replace the powerset $\{0,1\}^L$ with multiple subsets providing a
smaller solution space. Classifier chain methods \cite{CCReview}
use the conditional probability chain rule to traverse parts of the
solution \cite{PCCInferenceSurvey}, seek optimal chain structures \cite{BeamSearch2}
or use fixed structures \cite{CT} rather than a fully-cascaded chain.

We can see that modelling sets of labels (rather than single labels)
leads to increased classification accuracy, but it also leads to
greater complexity. Therefore multilabel classification should only be
used when there is dependence between labels to exploit. If we find
that from a set of 20 labels (with solution space $2^{20} = 1048576$)
the first 10 labels are independent of the second 10, the data can
be split into two 10 label problems (with solution space $2\times 2^{10}
= 2048$), which maintains the possible classification accuracy, but reduces the
solution space by a factor of 512. Therefore, before performing
multilabel classification, the data labels should be examined and
dependencies be identified.

But what exactly are we measuring to determine if there is dependence
between labels? 
Two random variables are considered Independent if
\begin{align*}
  P(Y_i, Y_j) = P(Y_i)P(Y_j)
\end{align*}
Assuming that each label $y_i$ is a Bernoulli random
variable $Y_i$ with parameter $p_i$ (such that $P(Y_i = 1) = p_i$),
the independence equation becomes
\begin{align*}
  p_{ij} = p_ip_j
\end{align*}
where $p_{ij} = P(Y_i = 1, Y_j = 1)$. The covariance between two
Bernoulli random variables is
\begin{align*}
  \rho_{ij} &= \mathbb{E}[Y_iY_j] - \mathbb{E}[Y_i]\mathbb{E}[Y_j] \\
  &= p_{ij} - p_ip_j
\end{align*}
Therefore, if the covariance is zero the random variables are
independent, if the covariance is not zero, the random variables are
not independent. For multilabel data, all labels should have
covariance, simply due to them all being dependent on
$\vec{x}$. Therefore the independence between labels can be measured
by the labels' conditional covariance (the label covariance
conditioned on $\vec{x}$). Therefore, to determine if a pair of labels
are dependent we are required to model
\begin{align*}
  \rho_{ij} = f(\vec{x})
\end{align*}
If we find $f(\vec{x}) = 0$, then the labels are independent. If
$f(\vec{x})$ is constant, then the conditional covariance is independent
of $\vec{x}$. otherwise it is dependent on $\vec{x}$.

Analytical models have been proposed to measure conditional covariance
between labels \cite{liu2019copula,gonccalves2015multi}, but they only
examined constant conditional covariance (the conditional covariance
might depend on $\vec{x}$) and so the accuracy of their measurements
is unclear. In the following sections, we present three analytical
models for measuring the conditional covariance and propose a simple
experiment for assessing the accuracy of the conditional covariance
measurements.

\section{Estimating label covariance using a Probit model}

\label{sec:probit}

The recorded observations of binary label $y_i$ in multilabel data can be
considered a sample from a Bernoulli random variable with parameter $p_i$.
Dependence between Bernoulli random variables can be measured using
their covariance $\rho_{ij}$. In this section we examine the Normal copula and
show how to measure label covariance using a multivariate Probit model.

\subsection{Modelling label covariance using a Normal Copula}

If two Bernoulli variables $Y_i$ and $Y_j$ with proportions $p_i$ and
$p_j$ respectively, have covariance $\bcov$, we can write
\begin{align*}
  Y_i = \llbracket p_i > U_i \rrbracket \qquad
  Y_j = \llbracket p_j > U_j \rrbracket
\end{align*}
where $p_i$ and $p_j$ are the known Bernoulli proportions, $U_i$ and $U_j$
are $[0,1]$ Uniform random variables with covariance, and
$\llbracket x \rrbracket$ is 1 if $x$ is true, otherwise it is 0.
We can combine the two equations to
form a two dimensional Uniform distribution.
\begin{align*}
  \left [
  \begin{array}{c}
    Y_i \\
    Y_j
  \end{array} \right ]
  &=
  \left \llbracket
  \left [ 
  \begin{array}{c}
    p_i \\
    p_j
  \end{array} \right ]
  > 
  \left [
  \begin{array}{c}
    U_i \\
    U_j
  \end{array} \right ]
  \right \rrbracket
  \\
  \left [
  \begin{array}{c}
    U_i \\
    U_j
  \end{array} \right ]
  &\sim \unift(0,1,\ncov)
\end{align*}
where $\llbracket \vec{x} \rrbracket$ provides a vector of 1 and 0, 1
for each element of $\vec{x}$ that is true and 0 for each element that
is false, and $\unift(0,1,\ncov)$ is a two dimensional Uniform
distribution with domain $[0,1]^2$ and covariance $\ncov$ (chosen so
that the covariance between $Y_i$ and $Y_j$ is $\bcov$).

\begin{figure*}[t]
\includegraphics[width=\linewidth]{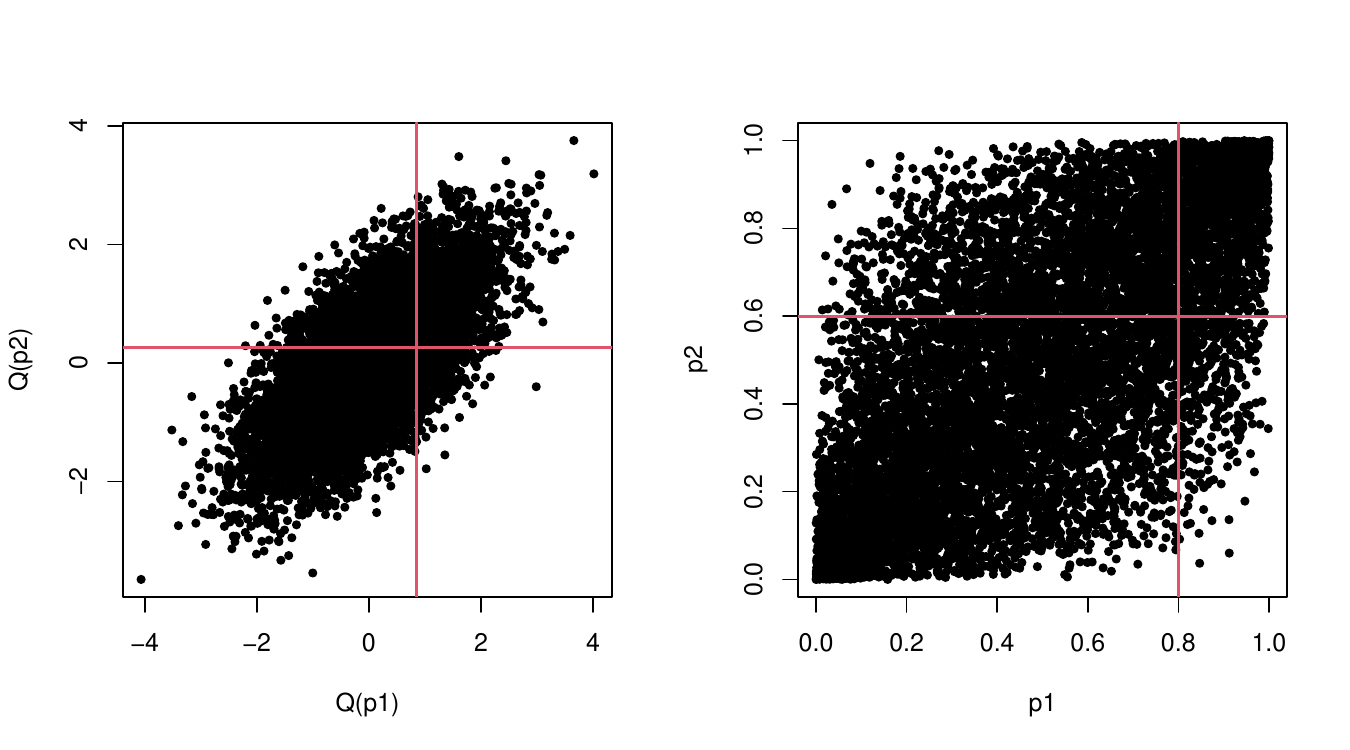}
\caption{A sample from a two dimensional correlated Normal distribution (left)
    and its conversion to a two dimensional correlated Uniform
    distribution. The red lines show the proportions 0.8 and 0.6 and
    their equivalent values in Normal distribution.}
  \label{fig:normalcop}
\end{figure*}

A simple method of sampling from a two dimensional correlated Uniform
distribution is to use a Normal copula \cite{leisch1998generation}. The process is to sample from a two dimensional Normal with mean
(0,0), variance (1,1) and covariance set to the desired amount
(between -1 and 1). Once the sample is obtained from the Normal
distribution, the quantiles of the sample will provide a sample from
the desired Uniform distribution. An example sample from a Normal and
the conversion to Uniform is shown in Figure \ref{fig:normalcop}.
The marginal distributions of this sample are still Uniform with range
$[0,1]$, but the resulting sample will be correlated.

Figure \ref{fig:normalcop} also shows red lines representing example
values of $p_i$ (0.8) and $p_j$ (0.6). The red lines divide the sample
space into four quadrants, each representing the joint probabilities
$P(Y_i = 1, Y_j = 1)$, $P(Y_i = 0, Y_j = 1)$, $P(Y_i = 1, Y_j = 0)$
and $P(Y_i = 0, Y_j = 0)$. The probability $P(Y_i = 1, Y_j = 1)$ can
be measured from the Uniform distribution as $P(U_i < p_i, U_j <
p_j)$. Since the correlated Uniform distribution is derived from a
Normal distribution, we can write the joint Bernoulli probability
as
\begin{align*}
P(Y_i = 1, Y_j = 1) = P(N_i < \probit(p_i), N_j < \probit(p_j))
\end{align*}
where $\probit(p)$ is the standard Normal quantile function (Probit), and
$N_i$ and $N_j$ are standard Normal with covariance $\ncov$.
By shifting the mean of the Normal distributions, we obtain
\begin{align*}
P(Y_i = 1, Y_j = 1) = P(N_i > 0 , N_j > 0)
\end{align*}
where the mean of $N_i$ and $N_j$ is $-\!\probit(p_i)$ and $-\!\probit(p_j)$ respectively.
Therefore, we
are able to compute $P(Y_i = 1, Y_j = 1)$ when given $p_i$, $p_j$ and
$\ncov$. Note that $P(Y_i = 1) = p_i$ and $P(Y_j = 1) = p_j$,
therefore the complete joint distribution
$P(Y_i = \{0,1\}, Y_j = \{0,1\})$ can be computed.
Given the proportions $p_i$ and $p_j$ and the covariance of
the Normal density $\ncov$, the complete set of joint
probabilities of $Y_i$ and $Y_j$ are defined. The joint Bernoulli
random variables are coupled to the joint Normal density, providing a copula.

\subsection{Multivariate Probit Model}

It was shown that joint probability of a pair of labels can be
represented using a two dimensional Normal distribution (as seen in
Figure \ref{fig:normalcop}).
In this section, we will describe how to associate the model
parameters to the data covariates $\vec{x}$.

We can write the joint label probability
in terms of a
two-dimensional Normal density function as
\begin{align}
  \label{eq:jointNorm}
  p_{ij} &= P\left (
  \left [
  \begin{array}{c}
    N_i \\
    N_j
  \end{array} \right ]
  > 
  \left [
  \begin{array}{c}
    0 \\
    0
  \end{array} \right ]\right)
\end{align}
where
\begin{align*}
  \left [
  \begin{array}{c}
    N_i \\
    N_j
  \end{array} \right ]
  &\sim\norm\left (\mu =
    \left [
  \begin{array}{c}
    \probit(p_i) \\
    \probit(p_j)
  \end{array} \right ], \Sigma =
  \left [ 
  \begin{array}{cc}
    1 & \ncov \\
    \ncov & 1
  \end{array} \right ]\right )
\end{align*}
and
$p_{ij} = P(Y_i = 1, Y_j = 1)$, 
$\norm$ is Normally distributed with mean equal to the $p_i$ and $p_j$
quantiles.
The joint Bernoulli probabilities are
simply each of the four quadrants of the Real sample space.
\begin{align*}
  p_{\bar{i}\bar{j}} &= P\left (
  \left [
  \begin{array}{c}
    N_i \\
    N_j
  \end{array} \right ]
  <
  \left [
  \begin{array}{c}
    0 \\
    0
  \end{array} \right ]\right) \\
  p_{{i}\bar{j}} &= P\left (
  \left [
  \begin{array}{c}
    N_i \\
    N_j
  \end{array} \right ]
  \begin{array}{c}
    > \\
    <
  \end{array}
  \left [
  \begin{array}{c}
    0 \\
    0
  \end{array} \right ]\right) \\
  p_{\bar{i}{j}} &= P\left (
  \left [
  \begin{array}{c}
    N_i \\
    N_j
  \end{array} \right ]
  \begin{array}{c}
    < \\
    >
  \end{array}
  \left [
  \begin{array}{c}
    0 \\
    0
  \end{array} \right ]\right)
\end{align*}
where $p_{\bar{i}\bar{j}} = P(Y_i = 0, Y_j = 0)$.
By expressing each of the model parameters in terms of the set of
covariates $\vec{x}$
we obtain the multivariate Probit model \cite{ashford1970multi,prentice1988correlated,liu2019copula}.
\begin{align}
  \label{eq:mvp}
  \begin{gathered}
  \begin{array}{l}
    \probit(p_i) = \gamma_{0} + \vec{\gamma}\vec{x} + \epsilon_i \\
    \probit(p_j) = \delta_{0} + \vec{\delta}\vec{x} + \epsilon_j
  \end{array}
\end{gathered}
\end{align}
where
\begin{align}
  \begin{gathered}
\left [
  \begin{array}{c}
    \epsilon_i \\
    \epsilon_j
  \end{array} \right ] \sim
  \norm\left (\mu =
  \left [
  \begin{array}{c}
    0 \\
    0
  \end{array} \right ], \Sigma =
  \left [ 
  \begin{array}{cc}
    1 & \ncov \\
    \ncov & 1
  \end{array} \right ]\right )
\end{gathered}
\end{align}
and
\begin{align}
  \label{eq:probitcov}
  \begin{gathered}
\log\left(\frac{1 + \ncov}{1 - \ncov}\right ) = \beta_{0} + \vec{\beta}\vec{x}
\end{gathered}
\end{align}
When fit to data, 
the multivariate Probit model estimates the coefficients
$\gamma_{0}$ and $\vec{\gamma}$ for $p_{i}$,
the coefficients $\delta_{0}$ and $\vec{\delta}$ for $p_{j}$,
and the coefficients $\beta_{0}$ and $\vec{\beta}$ for
$\ncov$ for the given data. Therefore, equation \eqref{eq:probitcov}
provides us with the relationship between the label conditional
covariance and $\vec{x}$.
The fitted coefficient $\beta_{0}$ measures the level of constant
covariance and $\vec{\beta}$ measures the level of covariance dependent
on each element of $\vec{x}$. If a significance test shows that these
are insignificant, then there is no evidence of label covariance and
hence we assume no dependence between the labels.

Note that each of the coefficients that estimate $p_i$ and $p_j$ are
estimated using only the marginal distribution for each label $Y_i$
and $Y_j$, and so are not influenced by the estimation of the
covariance $\ncov$. So even though the model estimates all
parameters at once, the estimate $\ncov$ is of the conditional
covariance, used to assess conditional dependence of the labels.
Note as well that the estimate of the conditional covariance $\ncov$ is the
covariance of the joint Normal distribution. The label conditional
covariance $\bcov$ (measuring independence) can be calculated as
$p_{ij} - p_ip_j$.

\subsection{Normal Copula Problem}

\label{sec:normprob}

A problem with the Normal copula representation comes from the
dependence between the mean and covariance matrix, which we will now
explain. If we hold the Normal covariance matrix constant, but vary
the mean, the proportion of the multilabel labels change and also the
covariance of the multilabel labels changes.

The Normal copula has the benefit of having less restrictive
parameters when compared to the multilabel proportions and joint
proportions,
due to the change in domain from $[0,1]^L$ to $\mathbb{R}^L$.
The mean
vector can be set to any real vector, which is then mapped to the
$[0,1]$ domain for the binary multilabel proportions. The Normal
covariance matrix requires a diagonal of 1 and all off-diagonal values
to be in the domain $[-1,1]$. Therefore, we can model each mean with a
function $\mu_i = f(\vec{x}) \in \mathbb{R}^L$, the associated label
proportion is $p_i = \Phi(\mu_i) \in [0,1]$, where $\Phi()$ is the
Normal cumulative probability function. Assuming that the multilabel
proportions $p_i$ and hence $\mu_i$ are dependent on $\vec{x}$, a
simple function for the mean is the linear function:
\begin{align*}
  \mu_{i} = \vec{\beta_i}\vec{x} + \beta_{i0}
\end{align*}
where $\vec{\beta_i}$ and $\beta_{i0}$ are the model coefficients (to be
fitted to data).  This linear function provides coefficients that allow
us to determine the effect of each $x_i$ on $\mu_i$ and hence the proportion
$p_i$. Other functions can also be used (for example, $f(\vec{x})$
could be a feed forward network), but it should be chosen to suit
the data.  The covariance values $\tau_{ij}$ are limited to the domain
$[-1,1]$, therefore we can model the real value
$\log\left (\frac{1+\tau_{ij}}{1-\tau_{ij}}\right )$ using an
unrestricted function. Again, a simple linear function is:
\begin{align*}
  \log\left (\frac{1+\tau_{ij}}{1-\tau_{ij}}\right ) = \vec{\gamma_{ij}}\vec{x} + \gamma_{ij0}
\end{align*}
where $\vec{\gamma_{ij}}$ and $\gamma_{ij0}$ are the model coefficients (to be
fitted to data) and again more complex functions can be used if
appropriate.

The coefficients of these models $\vec{\beta_i}$, $\beta_{i0}$,
$\vec{\gamma_{ij}}$ and $\gamma_{ij0}$ provide us insight as to how
the observation vector $\vec{x}$ effects the mean and covariance of
the Normal copula and hence seem to also describe the effect on the
proportions and covariance of the multilabel model labels. But even
though the proportions $p_i$ are a function of $\mu_i$, the binary
covariance $\rho_{ij}$ is a function of each of $\mu_i$, $\mu_j$
and $\tau_{ij}$, therefore the coefficients $\vec{\gamma_{ij}}$ do not
tell the full story. We will now examine the dependence of 
$p_{ij}$ and $\rho_{ij}$ (the Bernoulli parameters) on $\mu_i$,
$\mu_j$ and $\tau_{ij}$ (the Normal parameters).

If we have the copula model:
\begin{align*}
  \Phi^{-1}(p_1) = \mu_1 &= \beta_{10} + \beta_{11} x_1 + \beta_{12} x_2 \\
  \Phi^{-1}(p_2) = \mu_2 &= \beta_{20} \\
  \log\left (\frac{1+\tau_{12}}{1-\tau_{12}}\right ) &= \gamma_0
\end{align*}
so $p_1$ is dependent on $\vec{x}$, but $p_2$ and $\tau_{12}$ are
constant (independent of $\vec{x}$).  The coefficients $\beta_{11}$ and $\beta_{12}$ show how the
proportion $p_1$ reacts when $x_1$ and $x_2$ change. The coefficient
$\beta_{20}$ shows that the second label has constant probability of
being 1 and is independent of $\vec{x}$. The coefficient $\gamma_0$
shows that $\tau_{12}$ is constant and independent on $\vec{x}$, but
what does it say about the label covariance $\rho_{12}$ or the joint
probability $p_{12}$? If $x_1$ increases to $x_1 + \delta_x$, $p_1$
will increase to $p_1 + \delta_p$, but $p_2$ and $\tau_{12}$ will
remain constant. The joint probability $p_{12}$ (of $Y_1$ and $Y_2$) can be written as:
\begin{align*}
  p_{12} &= \int_0^\infty \int_0^\infty f(x_1, x_2; \vec{\mu}_{12}, \Sigma_{12}) dx_1dx_2 \\
  &= \int_{-\mu_1}^\infty \int_{-\mu_2}^\infty f(x_1, x_2; \vec{0}, \Sigma_{12}) dx_1dx_2 \\
  &= P(X_1 > -\mu_1, X_2 > -\mu_2; \vec{0}, \Sigma_{12}) 
\end{align*}
where $f$ is the multivariate Normal density.
An increase in $x_1$ by $\delta_x$ results in an increase in $\mu_1$ by $\delta_{\mu} = \beta_{11}\delta_x$.
\begin{align*}
  p_{12} + \delta   &= P(X_1 > -\mu_1 - \delta_{\mu}, X_2 > -\mu_2; \vec{0}, \Sigma_{12}) 
\end{align*}
So $\delta$ the change in $p_{12}$ resulting from an increase in $x_1$ is
\begin{align*}
  \delta &= P(-\mu_1 - \delta_{\mu} < X_1 < -\mu_1, X_2 > -\mu_2; \vec{0}, \Sigma_{12}) \\
   &= P(\mu_1  < X_1 < \mu_1 + \delta_{\mu}, X_2 < \mu_2; \vec{0}, \Sigma_{12})
\end{align*}
The proportion $p_1$ is:
\begin{align*}
  p_{1} &= P(X_1 < \mu_1; 0, 1) 
\end{align*}
and the change in $p_1$ due to a change in $x_1$ is
\begin{align*}
  \delta_p &= P(\mu_1 < X_1 < \mu_1 + \delta_{\mu}; 0, 1) 
\end{align*}
The covariance, and the change in covariance resulting from an increase in $x_1$ is:
\begin{align*}
  \rho_{12} &= p_{12} - p_1p_2 \\
  \rho_{12} + \delta_{\rho}  &= (p_{12} + \delta) - (p_1 + \delta_p)p_2 \\
            &= (p_{12} - p_1p_2) + \delta -  \delta_p p_2 \\
  \delta_{\rho} &= \delta -  \delta_p p_2
\end{align*}
The product $\delta_p p_2$ can be shown as a two-dimensional Normal with no covariance:
\begin{align*}
  \delta_p p_2 &= P(\mu_1 < X_1 < \mu_1 + \delta_{\mu}, X_2 < \mu_2; \vec{0}, I) 
\end{align*}
so the change in covariance between $p_1$ and $p_2$ due to a change in $x_1$ is
\begin{multline}
  \delta_{\rho} = 
  P(\mu_1  < X_1 < \mu_1 + \delta_{\mu}, X_2 < \mu_2; \vec{0}, \Sigma_{12}) \\
  - P(\mu_1 < X_1 < \mu_1 + \delta_{\mu}, X_2 < \mu_2; \vec{0}, I)
\end{multline}
The change in covariance $\delta_{\rho}$ is zero if $\Sigma_{12} = I$
(if the copula covariance matrix is the identity matrix).  Otherwise
the label covariance changes if $x_1$ changes.  The copula covariance
matrix $\Sigma_{12} = I$ only if $\tau_{12} = 0$. 
If $\Sigma_{12} = I$, then
\begin{align*}
  p_{12} &= P(X_1 > -\mu_1, X_2 > -\mu_2; \vec{0}, I) \\
         &= P(X_1 > -\mu_1; 0, 1)P(X_2 > -\mu_2; 0, 1) \\
         &= P(X_1 < \mu_1; 0, 1)P(X_2 < \mu_2; 0, 1) \\
         &= p_1p_2
\end{align*}
and so $\rho_{12} = 0$, implying no covariance between the pair of labels.

This analysis has shown that 
we can use the copula covariance coefficients to determine
if there is covariance between labels, but we are unable to use the
copula covariance coefficients to determine if the covariance is
constant or to provide insight into the dependence of the covariance
on $\vec{x}$. We are unable to use the copula method to determine if $\rho_{ij}$ is constant or
if $\rho_{ij}$ is dependent on $\vec{x}$.

\begin{figure}[t]
\begin{center}
    \includegraphics[width=0.49\linewidth]{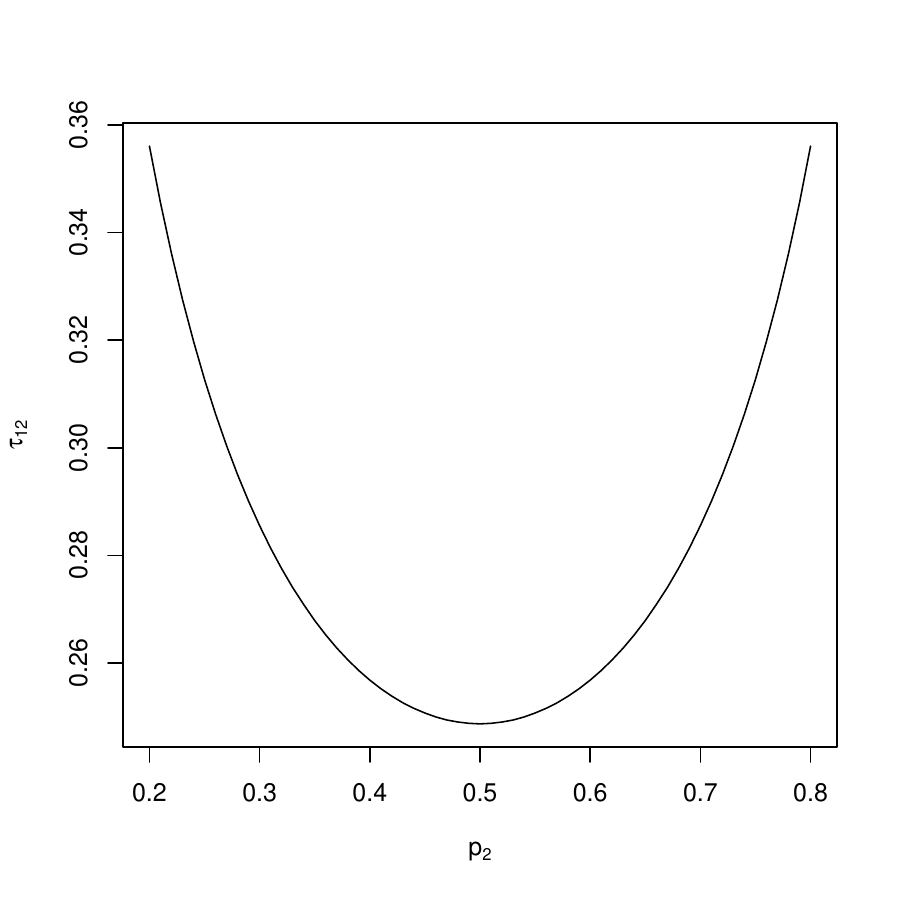}\hfill
    \includegraphics[width=0.49\linewidth]{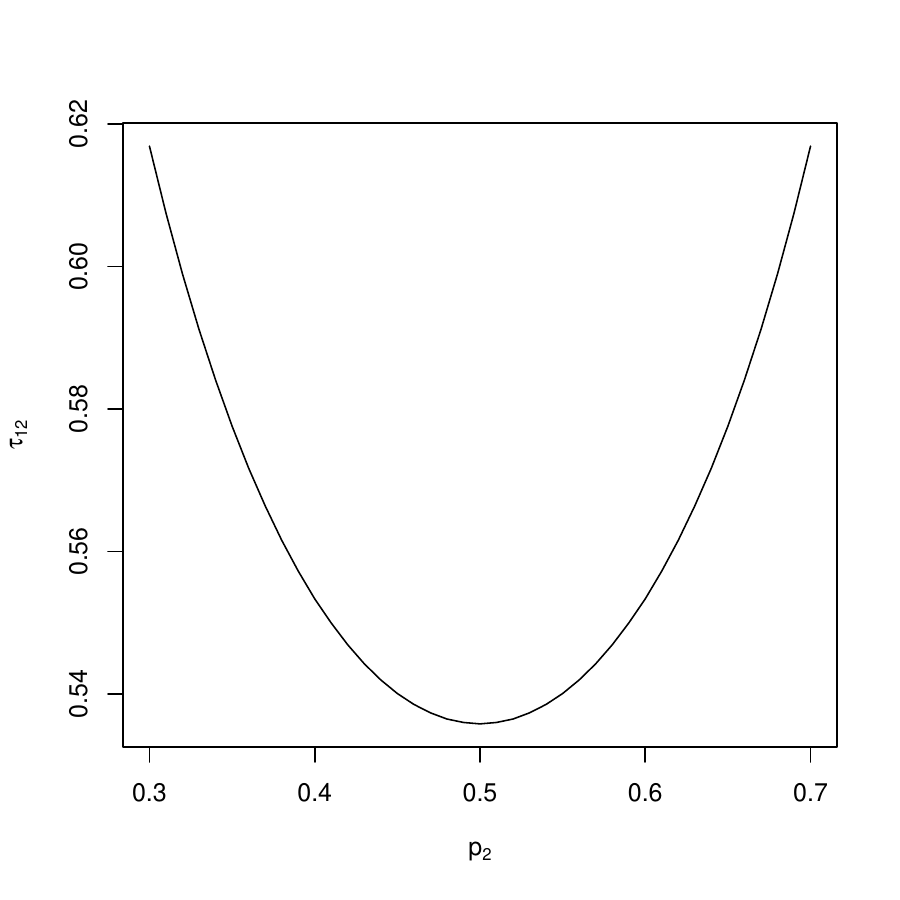}
  \end{center}
\caption{The effect of changing $p_2$ on the Probit covariance
    $\tau_{ij}$ when $p_1 = 0.5$ and the binary covariance
    $\rho_{ij} = 0.04$ (left) and $\rho_{ij} = 0.09$ (right).}
  \label{fig:tau_rho}
\end{figure}

The problem can also be shown by holding $\rho_{ij}$ constant and varying
$p_i$.
Figure \ref{fig:tau_rho} shows the relationship between $\tau_{ij}$ and
$p_i$ when $\rho_{ij}$ is held constant. The dependence of $\tau_{ij}$
on $p_i$ demonstrates that $\tau_{ij}$ will show dependence on the
conditional variables of $p_i$, being $\vec{x}$, even when $\rho_{ij}$ is constant.
This implies that the Probit model can be used to determine if
conditional covariance between labels exist, but it cannot be used to
identify if the conditional covariance is dependent on the
observations $\vec{x}$.

\section{Alternative Models}

\label{sec:anternative}

We have shown that the multivariate Probit model shows conditional
covariance when it does not exist. In this section, we present two
alternative models for measuring label conditional covariance.

\subsection{Staged Logistic Regression}
\label{sec:ucd}

Conditional covariance can be measured using the joint and marginal
conditional probabilities.
\begin{align}
  \label{eq:indepcondrho}
  \bcov(\vec{x}) = P(Y_i, Y_j|\vec{x}) - P(Y_i|\vec{x})P(Y_j|\vec{x})
\end{align}
From this model, we can easily model the joint ($P(Y_i, Y_j|\vec{x})$)
and marginal ($P(Y_i|\vec{x})$ and $P(Y_j|\vec{x})$) label
probabilities using a logistic regression, whose model coefficients
will show the dependence of the probabilities on $\vec{x}$. But we
want a model for $\bcov(\vec{x})$; combining the probability models
will not provide us with coefficients that measure the conditional covariance.

We can reduce the number of models to one by observing that we can
model the conditional covariance as the difference between $p_{ij}$
and $p_ip_j$.
Using the estimate $p_ip_j$ as a baseline, we can reveal that $Y_i$ and
$Y_j$ are not independent if $p_{ij}$ departs from the baseline.
Therefore, we can examine the effect of each $\vec{x}$ on the
dependence of $Y_i$ and $Y_j$ by modelling the
joint probability, using the independent estimate as a baseline.
\begin{align}
  \label{eq:dep1}
  p_{ij} = \sigma\left(\beta_0 + \sum_{k=1}^n \beta_kx_k\right) + p_ip_j
\end{align}
where $\sigma()$ is the sigmoid function, $x_k$ are the elements of
$\vec{x}$, and $\beta_k$ are the model coefficients.
If the coefficients $\beta_k$ are zero (or not statistically significant
from zero), then there is no evidence that $Y_i$ and $Y_j$ are not
independent. If the coefficient $\beta_0$ is non-zero, then $Y_i$ and
$Y_j$ are not independent regardless of each $x_k$. If $\beta_k$ is
non-zero, then the independence of $Y_i$ and $Y_j$ is effected by $x_k$.

After observing \eqref{eq:dep1}, it might seem that we can model the
difference $p_{ij} - p_ip_j$, but remember that we do not know
$p_{ij}$, we only have a sample from the Bernoulli distribution with
parameter $p_{ij}$. A final adjustment to the model is needed to allow
for this; adding $p_ip_j$ to the model is also difficult to perform,
so instead, we model
\begin{align}
  \label{eq:dep2}
  \logit(p_{ij}) = \beta_0 + \sum_{k=1}^n \beta_kx_k + \logit(p_ip_j)
\end{align}
making $\logit(p_ip_j)$ an offset to the linear portion of
the model. This will provide different estimates for $\beta_k$, but
they still retain the property that if they are zero, then the
associated $x_k$ has an effect on the dependence between $Y_i$ and
$Y_j$. This provides the Staged Logit model
\begin{align}
  \label{eq:twostage}
  \text{Stage 1:} & \quad \logit(p_{i}) = \gamma_0 + \sum_{k=1}^n \gamma_kx_k \\
  & \quad \logit(p_{j}) = \delta_0 + \sum_{k=1}^n \delta_kx_k \\
    \text{Stage 2:} & \quad \logit(p_{ij}) = \beta_0 + \sum_{k=1}^n \beta_kx_k + \logit(p_ip_j)
\end{align}
where stage 1 models the effect of each covariate $x_k$ on the
probability of labels $y_i$ and $y_j$ ($p_i$ and $p_j$) and
stage two models the effect of each covariate $x_k$ on the
joint probability of labels $y_i$ and $y_j$ over the baseline
probability $p_ip_j$. Therefore the coefficients $\beta_k$
identify the size of the effect of $x_k$ on the dependence between $y_i$ and $y_j$.
Note if $\beta_0$ and $\beta_k$ are zero, we are left with
$\logit(p_{ij}) = \logit(p_ip_j)$, showing
conditional independence between the labels.

\subsection{Multivariate Bernoulli Model}

Given two binary labels $Y_i$ and $Y_j$ and their associated
covariates $\vec{x}$, we can provide a density function

\begin{align*}
P(Y_i = y_i, Y_j = y_j) = p_{00}^{(1-y_i)(1-y_j)}p_{01}^{(1-y_i)y_j}p_{10}^{y_i(1-y_j)}p_{11}^{y_iy_j}
\end{align*}
where $p_{lk}$ is the probability of $Y_i = l$ and $Y_j = k$ and
$p_{00} + p_{01} + p_{10} + p_{11} = 1$.
We can manipulate the function to gather terms.
\begin{align*}
\MoveEqLeft[3] P(Y_i = y_i, Y_j = y_j) & \\
  =& \exp\left[\log\left(p_{00}^{(1-y_i)(1-y_j)}p_{01}^{(1-y_i)y_j}p_{10}^{y_i(1-y_j)}p_{11}^{y_iy_j}\right)\right] \\
  &\exp\left[(1-y_i)(1-y_j)\log(p_{00}) + {(1-y_i)y_j}\log(p_{01}) + \right .\\
  &\qquad \left .{y_i(1-y_j)}\log(p_{10}) + {y_iy_j}\log(p_{11})\right] \\
                           =& \exp\left[\log(p_{00}) - y_i\log(p_{00}) - y_j\log(p_{00})  + \right .\\
                           &\qquad  y_iy_j\log(p_{00}) + y_j\log(p_{01}) - y_iy_j\log(p_{01}) + \\
   &\qquad\left . y_i\log(p_{10}) - y_iy_j\log(p_{10}) + y_iy_j\log(p_{11})\right] \\
                           =& \exp\left[\log(p_{00}) + y_i\log\left(\frac{p_{10}}{p_{00}}\right) + \right .\\ \qquad &\left .y_j\log\left(\frac{p_{01}}{p_{00}}\right) + y_iy_j\log\left(\frac{p_{00}p_{11}}{p_{01}p_{10}}\right)\right]
\end{align*}
If we let
\begin{align*}
  f_i &= \log\left(\frac{p_{10}}{p_{00}}\right ) \\
  f_j &= \log\left(\frac{p_{01}}{p_{00}}\right ) \\
  f_{ij} &= \log\left(\frac{p_{11}p_{00}}{p_{01}p_{10}}\right )
\end{align*}
the multivariate Bernoulli probability simplifies to
\begin{align*}
  P(Y_i = y_i, Y_j = y_j) &= \exp\left[\log(p_{00}) + y_if_i + y_jf_j + y_iy_jf_{ij}\right]
\end{align*}
There are four parameters in this model, but remember we have the
constraint that the joint Bernoulli probabilities must sum to
one. Therefore, $p_{00}$ is a function of $f_i$, $f_j$ and $f_{ij}$.
We can show this by representing the following expression in terms of
its probabilities.
\begin{align*}
  \MoveEqLeft[10] 1 + \exp(f_i) + \exp(f_j) + \exp(f_i + f_j + f_{ij}) \\
  &= 1 + \frac{p_{10}}{p_{00}} + \frac{p_{01}}{p_{00}} +
  \frac{p_{10}}{p_{00}}\frac{p_{01}}{p_{00}}\frac{p_{11}p_{00}}{p_{01}p_{10}} \\
  &= 1 + \frac{p_{10}}{p_{00}} + \frac{p_{01}}{p_{00}} +
    \frac{p_{11}}{p_{00}} \\
  &= \frac{p_{00} + p_{10} + p_{01} + p_{11}}{p_{00}} \\
  &= \frac{1}{p_{00}}
\end{align*}
since $p_{00} + p_{10} + p_{01} + p_{11} = 1$. Therefore:
\begin{align*}
  p_{00} & = \frac{1}{1 + \exp(f_i) + \exp(f_j) + \exp(f_i + f_j + f_{ij})} \\
  p_{10} & = \frac{\exp{(f_i)}}{1 + \exp(f_i) + \exp(f_j) + \exp(f_i + f_j + f_{ij})} \\
  p_{01} & = \frac{\exp{(f_j)}}{1 + \exp(f_i) + \exp(f_j) + \exp(f_i + f_j + f_{ij})} \\
  p_{11} & = \frac{\exp{(f_i + f_j + f_{ij})}}{1 + \exp(f_i) + \exp(f_j) + \exp(f_i + f_j + f_{ij})}
\end{align*}
The probability density function can be rewritten as
\begin{align*}
  \MoveEqLeft[4] P(Y_i = y_i, Y_j = y_j) \\
  &= \exp\left[\log(p_{00}) + y_if_i + y_jf_j + y_iy_jf_{ij}\right] \\
  &= p_{00}\exp(y_if_i + y_jf_j + y_iy_jf_{ij}) \\
  &= \frac{\exp(y_if_i + y_jf_j + y_iy_jf_{ij})}{1 + \exp(f_i) + \exp(f_j) + \exp(f_i + f_j + f_{ij})}
\end{align*}
This model can be expanded to model higher order label relationships \cite{46e24991-096c-3081-b1e4-6f18f3e8e540},
making it a generalised form of the Ising model
\cite{gonccalves2015multi}.

This density function has the unknowns $f_i$, $f_j$ and $f_{ij}$ that
can be estimated using maximum likelihood. Each of these parameters
are log of probability ratios and so are real valued.
We can also
introduce the model covariates $\vec{x}$ by making these parameters dependent
on them. For example, we can use a linear model of the covariates.
\begin{align}
  f_i(\vec{x}) = \beta_{i0} + \beta_{i1}x_1 + \ldots \beta_{im}x_m \\
  f_j(\vec{x}) = \beta_{j0} + \beta_{j1}x_1 + \ldots \beta_{jm}x_m \\
  \label{eq:fijb}
  f_{ij}(\vec{x}) = \beta_{ij0} + \beta_{ij1}x_1 + \ldots \beta_{ijm}x_m
\end{align}
In this case, the coefficients $\beta$ are estimated when performing
maximum likelihood.

Recall that if $p_{ij} = p_ip_j$, then the covariance is zero and the
labels are independent. Therefore, if the labels are independent, we find that
\begin{align*}
  \frac{p_{00}p_{11}}{p_{01}p_{10}}
  &= \frac{p_{\bar{i}}p_{\bar{j}}p_ip_j}{p_{\bar{i}}p_jp_ip_{\bar{j}}} = 1
\end{align*}
Therefore, if we fit the model and $f_{ij} = 0$, then we have shown
that the labels are independent. Since we will be modelling $f_{ij}$
as a function of the covariates $\vec{x}$, the coefficients of the
function will give us more insight into the independence between
the labels. If we use the function from equation \eqref{eq:fijb}, the
coefficient $\beta_{ij0}$ measures the constant covariance, while
the coefficient $\beta_{ijk}$ measures the covariance dependent on the
$k$th covariate. Therefore, to test for constant or dependent
covariance between a pair of labels, we simply test if the associated
model coefficient is significantly different from zero.

\section{Experiment}

\label{sec:experiment}

We have presented three models for measuring the conditional
covariance of multilabel data.
Previous research has shown the effectiveness of these models in
determining covariance between each of the labels, but it is unclear
if the covariance is dependent on the observations $\vec{x}$.
Each of the models provide a function of the form
\begin{align*}
f(\rho_{ij}) = \beta_0 + \beta_1x_1 + \beta_2x_2 + \ldots + \beta_mx_m
\end{align*}
We can use the coefficient $\beta_0$ to measure the constant covariance,
and the coefficients $\beta_i$ to measure the covariance dependent on $x_i$.
But they are not directly modelling $\rho_{ij}$, they are instead
modelling a function of $\rho_{ij}$. We want to examine the effect of
not modelling $\rho_{ij}$ directly.

In this section, we describe a set of experiments to measure the
effectiveness of each of the three models in identifying constant
covariance (independent of the observations $\vec{x}$), and dependent
covariance (dependent on the observations $\vec{x}$). These experiments are
kept as simple as possible to allow us to control the experimental
variables and eliminate unwanted effects.
The experiment involves
\begin{enumerate}
\item generating data with the desired properties,
\item fitting the analytical models to the data, then
\item testing if the model coefficients exhibit the properties of
  the data.
\end{enumerate}
Each generated data set will consist of two
labels $y_i$ and $y_j$ and one covariate $x$. The data will consist of
two states ($x = 0$ and $x = 1$), each having their own label
probabilities for $y_i$ and $y_j$. A set of
label values (0 or 1) will be sampled from each of the two states to obtain the
data set.

\subsection{Generating the Data}

For a given pair of labels, there are four states (00, 01, 10,
11). The joint distribution of these two labels is defined by $p_{00}$, $p_{01}$,
$p_{10}$ and $p_{11}$. The marginal probabilities are
$p_{1.} = p_{10} + p_{11}$ and $p_{.1} = p_{01} + p_{11}$ and the
covariance between the labels is $\rho = p_{11} - p_{1.}p_{.1}$.
Note that we only need the marginals and covariance to define the joint distribution:
\begin{itemize}
\item $p_{11} = \rho + p_{1.}p_{.1}$
\item $p_{10} = p_{1.} - p_{11}$
\item $p_{01} = p_{.1} - p_{11}$
\item $p_{00} = 1 - p_{11} - p_{01} - p_{10}$
\end{itemize}
To create the experiment data, the parameters
$\alpha_0$, $\alpha_1$, $\gamma_0$, $\gamma_1$, $\beta_0$ and $\beta_1$
are chosen. These are used to define the marginal probabilities and covariance.
\begin{align*}
  p_{1.} &= \alpha_0 + \alpha_1 x \\
  p_{.1} &= \gamma_0 + \gamma_1 x \\
  \rho &= \beta_0 + \beta_1x
\end{align*}
To generate the data, we
\begin{enumerate}
\item Set the parameters $\alpha_0$, $\alpha_1$, $\gamma_0$,
  $\gamma_1$, $\beta_0$ and $\beta_1$ to obtain the desired $p_{1.}$,
  $p_{.1}$ and $\rho$,
\item Calculate the label joint distribution $p_{11}$, $p_{10}$,
  $p_{01}$ and $p_{00}$ for $x = 0$
\item Take a random sample of size 500 from the label joint distribution and
  record these binary values as the first half of the data with $x = 0$.
\item Calculate the label joint distribution $p_{11}$, $p_{10}$,
  $p_{01}$ and $p_{00}$ for $x = 1$
\item Take a random sample of size 500 from the label joint distribution and
  record these binary values as the second half of the data with $x = 1$.
\end{enumerate}
This provides a data set of 1000 elements, two labels $y_1$ and $y_2$
and one covariate $x$.

Note that due to the way the data was created, if $\alpha_1 = 0$, the
first label is independent of $x$, if $\gamma_1 = 0$, the
second label is independent of $x$, and if $\beta_1 = 0$, the
covariance is independent of $x$. If the parameters are nonzero, then
the associated probabilities/covariance is conditional based on $x$.

The data parameters are chosen to identify if the models can 
\begin{enumerate}
\item detect no covariance between labels (setting $\beta_0 = 0$ and
  $\beta_1 = 0$),
\item detect constant covariance (setting $\beta_0 \ne 0$ and
  $\beta_1 = 0$),
\item detect constant and dependant covariance (setting $\beta_0 \ne 0$ and
  $\beta_1 \ne 0$), and
\item detect dependant covariance only (setting $\beta_0 = 0$ and
  $\beta_1 \ne 0$).
\end{enumerate}
The configurations for each experiment is presented
in Table \ref{tab:config}. 

To clarify the process, we will provide an
example of generating data for the configuration Dep19. This
configuration has $p_{1.} = 0.5$ for state 0 and 1 ($\alpha_0 = 0.5$,
$\alpha_1 = 0$), $\rho = 0.01$ for state 0 and $\rho = 0.09$ for state
1 ($\beta_0 = 0.01$, $\beta_1 = 0.08$),
and generates 25 data sets each using
a combination of $p_{.1} =$ 0.3,
0.4, 0.5, 0.6, 0.7 for state 0 and state 1. For this example, we will use $p_{.1} = 0.3$
for state 0 and $p_{.1} = 0.5$ for state 1 ($\gamma_0 = 0.3$, $\gamma_1 =
0.2$).
Using these parameters we can deduce the complete
joint distribution of the pair of labels for state 0:
\begin{itemize}
\item $p_{11} = \rho + p_{1.}p_{.1} = 0.01 + 0.5\times 0.3 = 0.16$
\item $p_{10} = p_{1.} - p_{11} = 0.34$
\item $p_{01} = p_{.1} - p_{11} = 0.14$
\item $p_{00} = 1 - p_{11} - p_{01} - p_{10} = 0.36$
\end{itemize}
A random sample of size 500 is taken from this distribution to obtain
500 binary label pairs and the
observation $x = 0$ is associated to all 500 label pairs.
The same is repeated for state 1, where $p_{1.} = 0.5$, $p_{.1} = 0.5$, $\rho =
0.09$, where $x = 1$ is associated to all 500 label pairs. The two
samples are combined to obtain a sample of size 1000.

For each of the three analytical models (Multivariate Probit,
Multivariate Bernoulli and Staged Logit), we use $\beta_0 + \beta_1x$ as the model of $\rho$.
By using only values of $x = 0$ (for state 0) and $x = 1$ (for state
1) in the data, we ensure
that $\beta_0$ measures the covariance of state 0 (since $\rho =
\beta_0$ when $x = 0$) and $\beta_1$
measures the change in covariance from state 0 to state 1 (since
$\rho = \beta_0 + \beta_1$ when $x = 1$). Therefore,
we expect:
\begin{itemize}
\item $\beta_0$ is not significant when $\rho = 0$ for state 0
\item $\beta_0$ is significant when $\rho \ne 0$ for state 0
\item $\beta_1$ is not significant when there is no change in $\rho$
  between state 0 and 1
\item $\beta_1$ is significant when there is change in $\rho$
  between state 0 and 1
\end{itemize}
We expect both $\beta_0$ and $\beta_1$ to be zero for the
``Detecting no covariance'' experiment, $\beta_0$ to be nonzero and
$\beta_1$ to be zero for the ``Detecting constant covariance''
experiments, both $\beta_0$ and $\beta_1$ to be nonzero for the
``Detecting constant and dependant covariance'' experiments, and
$\beta_0$ to be zero and $\beta_1$ to be nonzero for the ``Detecting
dependant covariance only'' experiment.

Once the data is generated, the three models are fit to the data and
their coefficients $\beta_0$ and $\beta_1$ are tested for
significance.  To ensure consistency, 100 replicates of data for each
configuration were generated and models fit to each replicate. The
results report the proportion of coefficients that showed significance
from a Wald test.
The experiment results are shown in Tables \ref{tab:results1} for
$\beta_0$ and
\ref{tab:results2} for $\beta_1$. The code for the experiment is available from \texttt{anonymised}.

\begin{table*}[t]
\caption{The set of configurations used in experiments. Each
  configuration uses $p_1 = 0.5$ and $p_2$ chosen from 0.3, 0.4, 0.5,
  0.6, 0.7 for each state. A sample of size 500 is taken from each state. $x = 0$ is
  associated to state 0, $x = 1$ is associated to state 1.}
\begin{center}
\begin{tabularx}{0.8\linewidth}{lcZc}
  \toprule
  Name & \multicolumn{3}{c}{Experiment Configuration} \\
  \midrule
  \multicolumn{4}{l}{\em Detecting no covariance ($\rho = 0$)} \\
  \multirow{2}{*}{Zero} & State 0: $x=0$ & \multirow{2}{*}{$p_1 = 0.5$, $p_2 \in \{0.3, 0.4, 0.5, 0.6, 0.7\}$} & $\rho = 0$ \\
       & State 1: $x=1$ & & $\rho = 0$ \\[0.5em]
  \multicolumn{4}{l}{\em Detecting constant covariance ($\rho \ne 0$,
  is constant across state 0 and 1)} \\  
  \multirow{2}{*}{Const1} & State 0: $x=0$ &
                                             \multirow{6}{*}{
$p_1 = 0.5$, $p_2 \in \{0.3, 0.4, 0.5, 0.6, 0.7\}$} & $\rho = 0.01$ \\
         & State 1: $x=1$ & & $\rho = 0.01$ \\[0.2em]
  \multirow{2}{*}{Const4} & State 0: $x=0$ & & $\rho = 0.04$ \\
         & State 1: $x=1$ & & $\rho = 0.04$ \\[0.2em]
  \multirow{2}{*}{Const9} & State 0: $x=0$ & & $\rho = 0.09$ \\
       & State 1: $x=1$ & & $\rho = 0.09$ \\[0.5em]
  \multicolumn{4}{l}{\em Detecting constant and dependant covariance ($\rho \ne 0$,
  changes from state 0 to 1)} \\    
  \multirow{2}{*}{Dep41}  & State 0: $x=0$ & \multirow{6}{*}{$p_1 = 0.5$,
                                       $p_2 \in \{0.3, 0.4, 0.5, 0.6,
                                       0.7\}$} & $\rho = 0.04$ \\
         & State 1: $x=1$ & & $\rho = 0.01$ \\[0.2em]
  \multirow{2}{*}{Dep49}  & State 0: $x=0$ & & $\rho = 0.04$ \\
         & State 1: $x=1$ & & $\rho = 0.09$ \\[0.2em]
  \multirow{2}{*}{Dep19}  & State 0: $x=0$ & & $\rho = 0.01$ \\
       & State 1: $x=1$ & & $\rho = 0.09$ \\[0.5em]
  \multicolumn{4}{l}{\em Detecting dependant covariance only ($\rho
  = 0$ for state 0 and $\rho \ne 0$ for state 1)} \\      
  \multirow{2}{*}{Dep01}  & State 0: $x=0$ & \multirow{6}{*}{$p_1 = 0.5$,
                                       $p_2 \in \{0.3, 0.4, 0.5, 0.6,
                                       0.7\}$} & $\rho = 0.00$ \\
         & State 1: $x=1$ & & $\rho = 0.01$ \\[0.2em]
  \multirow{2}{*}{Dep04}  & State 0: $x=0$ & & $\rho = 0.00$ \\
         & State 1: $x=1$ & & $\rho = 0.04$ \\[0.2em]
  \multirow{2}{*}{Dep09}  & State 0: $x=0$ & & $\rho = 0.00$ \\
         & State 1: $x=1$ & & $\rho = 0.09$ \\
  \bottomrule
\end{tabularx}
\end{center}
\label{tab:config}
\end{table*}

\begin{table}[ht]
  \caption{Detecting constant label covariance ($\beta_0$). Each value is the
    proportion (and standard deviation in parentheses) of trials that
    concluded that the data contained a constant covariance between
    the pair of labels.}
  \centering
  \begin{tabularx}{0.9\linewidth}{lYYY}
    \toprule
    & Multivariate Probit & Multivariate Bernoulli & Staged Logit \\ 
    \midrule
    \multicolumn{4}{l}{\em Detecting no covariance (expect 0.05)} \\
    Zero & 0.0412 (0.0211) & 0.0488 (0.0247) & 0.0568 (0.0273) \\[0.5em] 
    \multicolumn{4}{l}{\em Detecting constant covariance (expect high)} \\    
    Const1 & 0.1548 (0.0276) & 0.1732 (0.0471) & 0.1400 (0.0289) \\ 
    Const4 & 0.9620 (0.0208) & 0.9628 (0.0219) & 0.9548 (0.0238) \\ 
    Const9 & 1.0000 (0.0000) & 1.0000 (0.0000) & 1.0000 (0.0000)
    \\[0.5em]
    \multicolumn{4}{l}{\em Detecting constant and dependant covariance (expect high)} \\    
    Dep41 & 0.9668 (0.0236) & 0.9636 (0.0229) & 0.9568 (0.0261) \\ 
    Dep49 & 0.9636 (0.0246) & 0.9632 (0.0250) & 0.9624 (0.0183) \\ 
    Dep19 & 0.1512 (0.0436) & 0.1540 (0.0318) & 0.1400 (0.0440)
    \\[0.5em]
    \multicolumn{4}{l}{\em Detecting dependant covariance only (expect 0.05)} \\    
    Dep01 & 0.0428 (0.0219) & 0.0528 (0.0242) & 0.0488 (0.0271) \\ 
    Dep04 & 0.0400 (0.0216) & 0.0488 (0.0167) & 0.0404 (0.0190) \\ 
    Dep09 & 0.0508 (0.0229) & 0.0492 (0.0214) & 0.0512 (0.0289) \\ 
    \bottomrule
  \end{tabularx}
  \label{tab:results1}
\end{table}
  
\subsection{Examining the Results}

Table \ref{tab:results1} contains the proportion of times that the
constant covariance coefficient $\beta_0$ was found to be significantly different from
zero, where significance was tested at the 5\% level.  Therefore, we
expect that the cases where the coefficient should be not be
significant to show as $0.05$ (the assigned false positive rate).  The
value in parentheses is the standard deviation of the proportion over
the various values of $p_2$ (from 0.3 to 0.7).  Constant covariance
should not be present in the ``Detecting no covariance'' and
``Detecting dependant covariance only'' results and we find that all
results are close to the expected $0.05$.
Constant covariance
should be present in the 
``Detecting constant covariance'' and
``Detecting constant and dependant covariance'' results. We find that
when the initial covariance is 0.04 or 0.09, then all proportions are
high (greater than 0.95). We also find that when the initial
covariance is 0.01 the proportions are low, likely due to being
difficult to detect. The results from all three models are similar.

\begin{table}[ht]
  \caption{Detecting dependent label covariance ($\beta_1$). Each value is the
    proportion (and standard deviation in parentheses) of trials that
    concluded that the data contained a covariance that was dependent
    on $\vec{x}$ between the pair of labels.}
  \centering
  \begin{tabularx}{0.9\linewidth}{lYYY}
    \toprule
    & Multivariate Probit & Multivariate Bernoulli & Staged Logit \\ 
    \midrule
    \multicolumn{4}{l}{\em Detecting no covariance (expect 0.05)} \\
    Zero & 0.0460 (0.0289) & 0.0492 (0.0250) & 0.0496 (0.0244)
    \\[0.5em]
    \multicolumn{4}{l}{\em Detecting constant covariance (expect 0.05)} \\
    Const1 & 0.0496 (0.0207) & 0.0544 (0.0202) & 0.0520 (0.0220) \\ 
    Const4 & 0.0604 (0.0215) & 0.0628 (0.0239) & 0.1264 (0.0930) \\ 
    Const9 & 0.1068 (0.0607) & 0.1356 (0.0938) & 0.2716 (0.2627)
    \\[0.5em]
    \multicolumn{4}{l}{\em Detecting constant and dependant covariance (expect high)} \\    
    Dep41 & 0.5248 (0.0668) & 0.5064 (0.0932) & 0.4920 (0.1787) \\ 
    Dep49 & 0.9260 (0.0612) & 0.9012 (0.0871) & 0.7800 (0.2814) \\ 
    Dep19 & 0.9992 (0.0028) & 0.9996 (0.0020) & 0.9920 (0.0200)
    \\[0.5em]
    \multicolumn{4}{l}{\em Detecting dependant covariance only (expect high)} \\    
    Dep01 & 0.0968 (0.0295) & 0.1040 (0.0357) & 0.1060 (0.0357) \\ 
    Dep04 & 0.7500 (0.0760) & 0.7592 (0.0642) & 0.7148 (0.1228) \\ 
    Dep09 & 1.0000 (0.0000) & 1.0000 (0.0000) & 0.9988 (0.0044) \\ 
    \bottomrule
  \end{tabularx}
\label{tab:results2}
\end{table}

Table \ref{tab:results2} contains the proportion of times that the
dependent covariance coefficient $\beta_1$ (measuring dependent covariance) was found to be significantly different from
zero, where significance was tested at the 5\% level.  Again, 
cases where the coefficient should not be
significant should show as $0.05$ (the assigned false positive rate).  The
value in parentheses is the standard deviation of the proportion over
the various values of $p_2$ (from 0.3 to 0.7).  Dependent covariance
should not be present in the 
``Detecting no covariance'' and
``Detecting constant covariance'' results. We find that the results
are as expected for the ``no covariance'' results, but all of the
methods show a detection that is dependent on the covariance level
for the ``constant covariance'' results. The error is largest for the
Staged Logit model and smallest for the Multivariate Probit model.

Dependent covariance should be present in the 
``Detecting dependant covariance only'' and 
``Detecting constant and dependant covariance'' data. We find
that dependent covariance is detected and that again it is dependent
on the strength of the covariance.

To summarise the results all three models behaved as expected for most
experiments. Detecting covariance depends on the covariance
strength, which makes sense, since we are testing if the coefficient
is different from zero and smaller coefficients are more difficult to
distinguish from zero. All models falsely detected dependent
covariance in the data where only constant covariance was present,
where the Multivariate Probit model had the least error.

\subsection{Experiments using real data}

The results using controlled generated data have been provided and you
might be wondering where to find the results using real data. This
research is investigating if the three analytical models are able to detect
conditional covariance; using the controlled experiments we were able
to control the conditional covariance properties of the data and
investigate if the models were able to detect it.

Unfortunately, we don't have measurements for conditional covariance
for real data. If we could measure it, this research would not be
needed (since we are investigating the effetiveness of models that
allow us to measure conditional covariance in multilabel data).
Any results provided on real data would show each models measurement
of conditional covariance, but there is no ground truth data to
determine its accuracy and so the results would not be useful.

We have deliberately left out providing any measurements of
conditional covariance for real data to reduce the temptation of the
reader referring to these real data results in their research, since we would be
uncertain of the reliability of their measurements.

\section{Model Analysis}

\label{sec:analysis}

Our experiments found that all models had difficulty when exposed to
data with constant covariance, showing that a false dependent covariance existed.
In this section, we will examine the models to determine why this
occurred. We will examine the case where 
$p_1 = 0.5$ and $\rho = 0.09$ for both state 0 and state 1, and $p_2$
is adjusted from $0.3$ in state 0 to $0.5$ in state 1. So for this
example both $p_1$ and $\rho$ are independent of $\vec{x}$ (they are constant), but $p_2$
is dependent on $\vec{x}$.

\subsection{Multivariate Bernoulli}

For state 0, $p_1 = 0.5$, $p_2 = 0.3$ and $\rho = 0.09$, so $p_{11} =
0.24$, $p_{10} = 0.26$, $p_{01} = 0.06$ and $p_{00} = 0.44$.
The covariance term $f_{ij}$ becomes
\begin{align*}
f_{ij} = \log{\left (\frac{0.24\times 0.44}{0.06\times 0.26}\right )} = 1.912
\end{align*}
For state 1, $p_1 = 0.5$, $p_2 = 0.5$ and $\rho = 0.09$, so $p_{11} =
0.34$, $p_{10} = 0.16$, $p_{01} = 0.16$ and $p_{00} = 0.34$.
The covariance term $f_{ij}$ becomes
\begin{align*}
f_{ij}(\vec{x}) = \log{\left (\frac{0.34\times 0.34}{0.16\times 0.16}\right )} = 1.507
\end{align*}
We see that even though $\rho$ is held constant, $f_{ij}$ has changed. This is
due to the Bernoulli model using the approximation:
\begin{align*}
  \log(\rho) &= \log(p_{11}p_{00} - p_{10}p_{01}) \\
             &\approx \log(p_{11}p_{00}) - \log(p_{10}p_{01}) = f_{ij}
\end{align*}

\subsection{Multivariate Probit}

For state 0, 
$p_1 = 0.5$, $p_2 = 0.3$ and $\rho = 0.09$, giving the fitted multivariate
Probit parameters
\begin{align*}
  \left [
  \begin{array}{c}
    \mu_1 \\ \mu_2
  \end{array}
  \right ]  
  &= \Phi^{-1}\left (\left [
  \begin{array}{c}
    0.5 \\ 0.3
  \end{array}
  \right ]\right )
  =
  \left [
  \begin{array}{c}
    0 \\ -0.524
  \end{array}
  \right ]  \\
\Sigma &=
  \left [
  \begin{array}{cc}
    1 & 0.6164 \\
    0.614 & 1
  \end{array}
  \right ]
\end{align*}
providing $p_{11} = 0.24$ and hence, $\rho = 0.09$.

For state 1, $p_1 = 0.5$, $p_2 = 0.5$ and $\rho = 0.09$.
so
the fitted parameters are
\begin{align*}
  \left [
  \begin{array}{c}
    \mu_1 \\ \mu_2
  \end{array}
  \right ]  
  &= \Phi^{-1}\left (\left [
  \begin{array}{c}
    0.5 \\ 0.5
  \end{array}
  \right ]\right )
  =
  \left [
  \begin{array}{c}
    0 \\ 0
  \end{array}
  \right ]  \\
\Sigma &=
  \left [
  \begin{array}{cc}
    1 & 0.5358 \\
    0.5358 & 1
  \end{array}
  \right ]
\end{align*}
providing $p_{11} = 0.34$ and hence, $\rho = 0.09$.

We can see that even though $\rho$ remains constant, the value of
$\tau$ (in $\Sigma$) has changed due to $p_2$ changing. Therefore, our model of
$\tau$ with respect to $\vec{x}$ will show a change in $\tau$ due to a
change in $\vec{x}$.

\subsection{Staged Logit}

Using the Two-stage Logit model, both $p_1$ and $p_2$ are modelled
using logistic regressions. The second stage models $p_{11}$ using
$p_{1}p_{2}$ as an offset.

For state 0:  $p_1 = 0.5$, $p_2 = 0.3$ and $\rho = 0.09$, so
$p_{11} = 0.24$
\begin{align*}
  \log{\left( \frac{p_{11}}{1-p_{11}} \right )} &= \rho + \log{\left( \frac{p_1p_2}{1-p_1p_2} \right )} \\
-1.1526 &= \rho + -1.7346
\end{align*}
giving $\rho = 0.582$

For state 1, $p_1 = 0.5$, $p_2 = 0.5$ and $\rho = 0.09$ (only changing $p_2$), so
$p_{11} = 0.34$
\begin{align*}
  \log{\left( \frac{p_{11}}{1-p_{11}} \right )} &= \rho + \log{\left( \frac{p_1p_2}{1-p_1p_2} \right )} \\
-0.6632 &= \rho + -1.0986
\end{align*}
giving $\rho = 0.435$.
This error comes from the Staged Logit model using the approximation
\begin{align*}
\logit(p_{11}) &= \logit(\rho + p_1p_2) \\
  &\approx \logit(\rho) + \logit(p_1p_2)
\end{align*}
So all of the models do not directly model $\rho$ and so their estimates
are effected by changes in $p_1$ and $p_2$.

\section{Conclusion}

Independence between multilabel data labels cannot be measured
directly from the label values due to their dependence on the set of
covariates $\vec{x}$. Previous research has shown that label
independence can be be measured by examining the conditional label
covariance using a multivariate Probit model. Unfortunately, the 
multivariate Probit model provides an estimate of the copula
covariance, and so might not be reliable in estimating constant
covariance and dependent covariance.

In this article, we presented three models (Multivariate Probit, Multivariate
Bernoulli and Staged Logit) for estimating the constant and dependent
multilabel conditional label covariance.  We provided an experiment
that allowed us to observe each model's measurement of conditional
covariance. We found that all models measure constant and dependent
covariance equally well, depending on the strength of the covariance,
but the models all falsely detect that dependent covariance is present
for data where constant covariance is present. Of the three models,
the Multivariate Probit model had the lowest error rate.

\bibliographystyle{IEEEtran}
\bibliography{p}
\end{document}